\documentclass[sigconf]{acmart}
\usepackage{multirow}
\usepackage{enumitem}

\copyrightyear{2023}
\acmYear{2023}
\setcopyright{acmlicensed}
\acmConference[MM '23] {Proceedings of the 31st ACM International Conference on Multimedia}{October 29--November 3, 2023}{Ottawa, ON, Canada.}
\acmBooktitle{Proceedings of the 31st ACM International Conference on Multimedia (MM '23), October 29--November 3, 2023, Ottawa, ON, Canada}
\acmPrice{15.00}
\acmISBN{979-8-4007-0108-5/23/10}
\acmDOI{10.1145/3581783.3612159}

\begin{document}

\title{HSVLT: Hierarchical Scale-Aware Vision-Language Transformer for Multi-Label Image Classification}

\author{Shuyi Ouyang}
\email{oysy@zju.edu.cn}
\affiliation{%
  \institution{Zhejiang University}
  \city{Hangzhou}
  \country{China}
}

\author{Hongyi Wang}
\email{whongyi@zju.edu.cn}
\affiliation{
  \institution{Zhejiang University}
  \city{Hangzhou}
  \country{China}
}

\author{Ziwei Niu}
\email{nzw@zju.edu.cn}
\affiliation{
  \institution{Zhejiang University}
  \city{Hangzhou}
  \country{China}
}

\author{Zhenjia Bai}
\email{22221168@zju.edu.cn}
\affiliation{
  \institution{Zhejiang University}
  \city{Hangzhou}
  \country{China}
}

\author{Shiao Xie}
\email{22160144@zju.edu.cn}
\affiliation{
  \institution{Zhejiang University}
  \city{Hangzhou}
  \country{China}
}

\author{Yingying Xu}
\email{cs_ying@zju.edu.cn}
\affiliation{
  \institution{Zhejiang Lab}
  \city{Hangzhou}
  \country{China}
}
\author{Ruofeng Tong}
\email{trf@zju.edu.cn}
\affiliation{
  \institution{Zhejiang University}
  \city{Hangzhou}
  \country{China}
}
\author{Yen-Wei Chen}
\authornote{Corresponding Authors: Lanfen Lin and Yen-Wei Chen.}
\email{chen@is.ritsumei.ac.jp}
\affiliation{
  \institution{Ritsumeikan University}
  \city{Kusatsu}
  \country{Japan}
}

\author{Lanfen Lin}
\authornotemark[1]
\email{llf@zju.edu.cn}
\affiliation{
  \institution{Zhejiang University}
  \city{Hangzhou}
  \country{China}
}

\renewcommand{\shortauthors}{Shuyi Ouyang et al.}

\begin{abstract}

  The task of multi-label image classification involves recognizing multiple objects within a single image. Considering both valuable semantic information contained in the labels and essential visual features presented in the image, tight visual-linguistic interactions play a vital role in improving classification performance. Moreover, given the potential variance in object size and appearance within a single image, attention to features of different scales can help to discover possible objects in the image. Recently, Transformer-based methods have achieved great success in multi-label image classification by leveraging the advantage of modeling long-range dependencies, but they have several limitations. Firstly, existing methods treat visual feature extraction and cross-modal fusion as separate steps, resulting in insufficient visual-linguistic alignment in the joint semantic space.  Additionally, they only extract visual features and perform cross-modal fusion at a single scale, neglecting objects with different characteristics. To address these issues, we propose a Hierarchical Scale-Aware Vision-Language Transformer (HSVLT) with two appealing designs: (1)~A hierarchical multi-scale architecture that involves a Cross-Scale Aggregation module, which leverages joint multi-modal features extracted from multiple scales to recognize objects of varying sizes and appearances in images. (2)~Interactive Visual-Linguistic Attention, a novel attention mechanism module that tightly integrates cross-modal interaction, enabling the joint updating of visual, linguistic and multi-modal features. We have evaluated our method on three benchmark datasets. The experimental results demonstrate that HSVLT surpasses state-of-the-art methods with lower computational cost. 
\end{abstract}




\maketitle

\begin{figure}[t!]
  \centering
  \includegraphics[width=\linewidth]{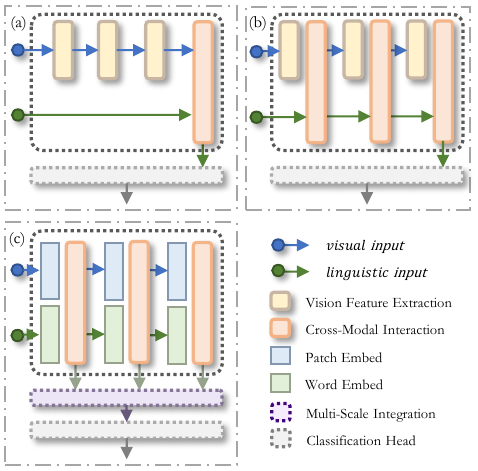}
  \caption{Comparison of existing Transformer-based architectures ((a) and (b)) for multi-label image classification with our HSVLT (c).}
  \label{fig1} \vspace{-15pt}
\end{figure}


\section{Introduction}

Multi-label image classification refers to the task of recognizing multiple objects within a single image. It yields great value for various applications such as image retrieval \cite{wei2019saliency}, human attribute recognition \cite{li2016human} and scene understanding \cite{shao2015deeply}.
Unlike single-label classification, the multi-label classification task 
presents a substantial challenge as it involves identifying multiple objects within a single image with imbalanced label distribution and varying categories, requiring abundant local visual information. 
The semantic information contained in the labels can help model the potential objects in the image, 
underscoring the importance of modeling global cross-modal relationships. 
As objects within the given image may differ significantly in size and appearance, extracting visual features and modeling visual-linguistic relationships at multiple scales can enhance the ability to capture different objects.  In addition, performing cross-scale aggregation help to leverage complementary  information from different scales for decision-making purposes.

Early methods for multi-label image classification \cite{wei2015hcp, yang2016exploit} primarily relied on object detection.
Subsequently, some methods \cite{chen2018recurrent, wang2017multi} attempted to model spatial correlation to enhance the interaction between image regions. 
As research progressed, scholars began exploring how to leverage correlations between images and labels to improve classification accuracy. The CNN-RNN framework \cite{wang2016cnn} and PLA \cite{yazici2020orderless} were proposed to implicitly  model label dependencies using RNN and sequentially predict labels.
Given the significance of global cross-modal modeling in the multi-label classification task, the advantage of Transformers \cite{vaswani2017attention} with attention mechanisms in long-range modeling makes it an ideal fit for this task. 
Consequently, methods based on Vision Transformer (ViT)~\cite{dosovitskiy2020image}  have demonstrated exceptional performance in the multi-label classification task. Figure~1(a) illustrates one of the transformer-based architectures for multi-label classification,  i.e., M3TR \cite{zhao2021m3tr}, which fuses visual features and linguistic features after visual feature extraction.
Another architecture, shown in Figure~1(b), is employed in TSFormer \cite{zhu2022two}. This architecture alternates visual feature extraction with cross-modal fusion.
Through modeling long-range dependencies, Transformer-based methods have achieved remarkable success, yet there is still potential for improvement.
Existing Transformer-based methods treat visual feature extraction and cross-modal fusion as two distinct processes, leading to insufficient visual-linguistic alignment in the joint semantic space. Moreover, they only perform visual feature extraction and cross-modal fusion at a single scale, which limits the acquisition of comprehensive multi-modal information.


After a thorough analysis of previous successful works and the requirements of the multi-label classification task, we contend that effective methods for this task should possess the following characteristics: (i) A strong cross-modal encoder to capture both local visual and global visual-linguistic information. 
Discriminating between objects with confusing appearances in a single image can pose difficulties, underscoring the tight visual-linguistic interactions to ensure accurate classification.
Nevertheless, efficient interaction between the two modalities remains a challenge. 
(ii)~Multi-scale information interaction to capture dependencies across multiple scales and resolve complex scale differences. 
The existing methods only focus on a single scale and fail to consider cross-modal features from a scale-aware perspective, resulting in the omission of certain objects.

Therefore, in light of the aforementioned analysis, 
we propose \textbf{H}ierarchical \textbf{S}cale-Aware \textbf{V}ision-\textbf{L}anguage \textbf{T}ransformer (HSVLT), a novel multi-label classification architecture as depicted in Figure~1(c). 
We introduce a hierarchical multi-scale architecture comprising multiple stages, which leverages Scale Transformation blocks and Channel Unification blocks to achieve decreasing spatial resolution of visual features and channel unification of linguistic features. HSVLT enables visual feature extraction and cross-modal interaction within multiple scales, while also aggregating multi-modal features extracted from different scales. 
At each scale, we design a novel attention mechanism module called Interactive Visual-Linguistic Attention (IVLA) that integrates cross-modal interaction. 
IVLA is a component of the joint vision-language encoder that can update visual, linguistic, and multi-modal features simultaneously.
By considering interactive cross-modal cues, IVLA obtains rich local visual features and tight global visual-linguistic relationships, which enhances visual-linguistic alignment in the joint semantic space and improves multi-label image classification accuracy.
Additionally, we propose a Cross-Scale Aggregation (CSA) module to model multi-scale information as a whole, taking into account complementary multi-modal information at different scales in a holistic manner.
The proposed multi-scale architecture acquires comprehensive multi-modal information from both within and across scales, addressing the issue of neglecting unconspicuous objects. 

\begin{figure*}[ht!]
    \centering
    \includegraphics[width=1.0\linewidth]{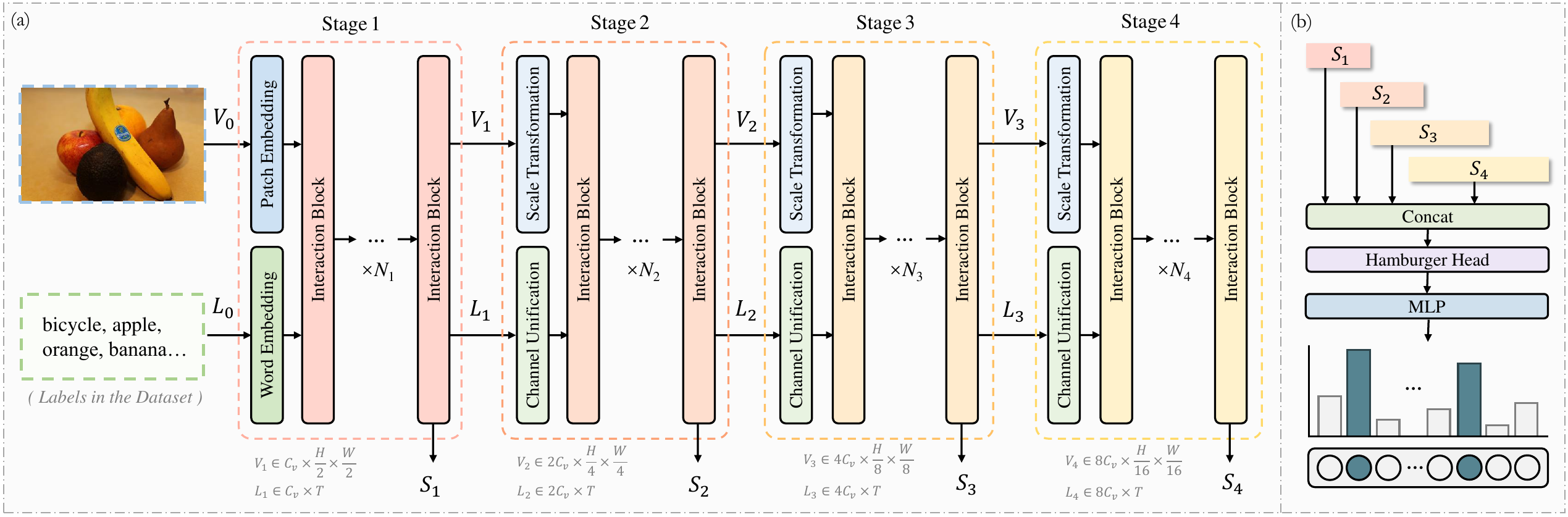}
    \caption{An illustration of HSVLT. The multi-scale joint vision-language encoder network is presented in (a). (b) shows the cross-scale aggregation module for multi-label classification.
    First, the input image $V_0$ and labels $L_0$ are sent to Joint Vision-Language Encoder. At the beginning of each stage, we down-sample visual features and unify the channel dimensions of visual and linguistic features. There are $N_i$ interaction blocks in $i$-th stage.
    Interaction blocks learn joint visual features $V_i, i \in \{ 1,2,3,4\}$, linguistic features $L_i, i \in \{ 1,2,3,4\}$ and multi-modal features $S_i, i \in \{ 1,2,3,4\}$, 
    which contains local visual details and global visual-linguistic cues.
    $S_1,S_2,S_3,S_4$ are sent to the cross-scale aggregation module (b) for multi-label classification prediction.
    }
    \label{fig2}
\end{figure*}

In summary, our contributions are four-folded:
\begin{enumerate} [itemsep=2pt,topsep=0pt,parsep=0pt]
    
    \item 
    We introduce a hierarchical multi-scale architecture with stages of decreasing spatial resolution. This enables visual feature extraction and cross-modal interaction within scales, as well as multi-modal feature aggregation across scales.

    \item 
    We propose Interactive Visual-Linguistic Attention (IVLA) module in our joint vision-language encoder. IVLA enables the joint updating of visual features, linguistic features and multi-modal features considering interactive cross-modal cues, which enhances visual-linguistic alignment in the joint semantic space.

    \item 
    We design Cross-Scale Aggregation (CSA) module to leverage complementary information from different scales for decision-making purposes. Taking into account the multi-modal information from different scales and capturing complementary knowledge across scales, CSA enhances the recognition of objects of varying sizes and appearances.

    \item 
    Building on these designs, we introduce HSVLT, a novel approach for multi-label classification. 
    We conducted thorough experiments on HSVLT with three benchmark datasets, showing that HSVLT outperforms current state-of-the-art methods with lower computational cost.
    
\end{enumerate}

\section{Related Works}
\paragraph{\textbf{Multi-Label Image Classification}} 
For multi-label image classification, early approaches \cite{wei2015hcp, yang2016exploit} were based on object detection, where objects in the image were detected and located before being individually classified.
Subsequently, researchers explored how to leverage label correlations to enhance classification accuracy.   Gong \emph{et al.} \cite{gong2013deep} employed specialized loss functions to optimize label correlations. Chen \emph{et al.} \cite{chen2019multi} utilized a directed graph to model label dependencies.
In addition, the effectiveness of exploring spatial dependencies with label semantics has also been demonstrated.
Wang \emph{et al.} \cite{wang2017multi} proposed the utilization of spatial transformer layers to emphasize the image regions relevant to the labels, while Chen \emph{et al.} \cite{chen2018recurrent} devised the recurrent attention reinforcement module for the same purpose.
Wu \emph{et al.} \cite{wu2021gm} reformulated the multi-label image classification problem as a graph matching structure, which incorporates instance space relations, label semantic relevance, and instance-label assignment probability into the framework.
Recently, Transformer-based methods have shown improved ability in modeling long-range cross-modal dependencies and made significant progress in multi-label classification.
In these methods, M3TR \cite{zhao2021m3tr} separately learns ternary relationships inter- and intra- modalities and performs semantic cross-attention, while Zhu \emph{et al.} designed Two-Stream Transformer \cite{zhu2022two} to extract global features and correlations of label semantics separately.  


\paragraph{\textbf{Vision Transformer}}
Transformer \cite{vaswani2017attention} models have been widely used for several computer vision tasks. The ViT model applies self-attention in shallow layers enhancing performance for vision tasks.
Recent Transformer-based models have achieved impressive results across a range of vision tasks, including image classification~\cite{dosovitskiy2020image,liu2021swin}, object detection~\cite{carion2020end}, and semantic segmentation~\cite{strudel2021segmenter}. 
The self-attention mechanism of the Transformer allows the models to effectively capture long-range dependencies in images, establishing global contextual information.
In addition, the multi-scale design of the Transformer \cite{zheng2021rethinking, gu2022multi} allows the model to aggregate features of different scales in the image, improving its overall performance in vision tasks.
For multi-label image classification task, various Transformer-based methods have been presented using advantages of long-range dependencies. 
Lanchantin \emph{et al.} \cite{lanchantin2021general} introduced a general framework based on a Transformer, with the use of a ternary coding scheme for training. Cheng \emph{et al.} \cite{cheng2022mltr} proposed a Multi-Label Transformer architecture comprising of window partitioning, pixel attention within windows, and cross-window attention. Zhao \emph{et al.} \cite{zhao2021m3tr} presented a multi-modal multi-label recognition transformer framework that includes ternary relationship learning for both inter- and intra-modal information. Zhu \emph{et al.} \cite{zhu2022two} introduced the Two-Stream Transformer, which utilizes spatial and semantic streams to respectively learn the visual perception and label semantics and their correlation.
However, in current Transformer-based approaches, visual feature extraction and cross-modal fusion are limited to a single scale, failing to capture information at diverse scales within an image, which negatively impact the classification performance. 


\begin{figure*}[ht!]
    \centering
    \includegraphics[width=1.0\linewidth]{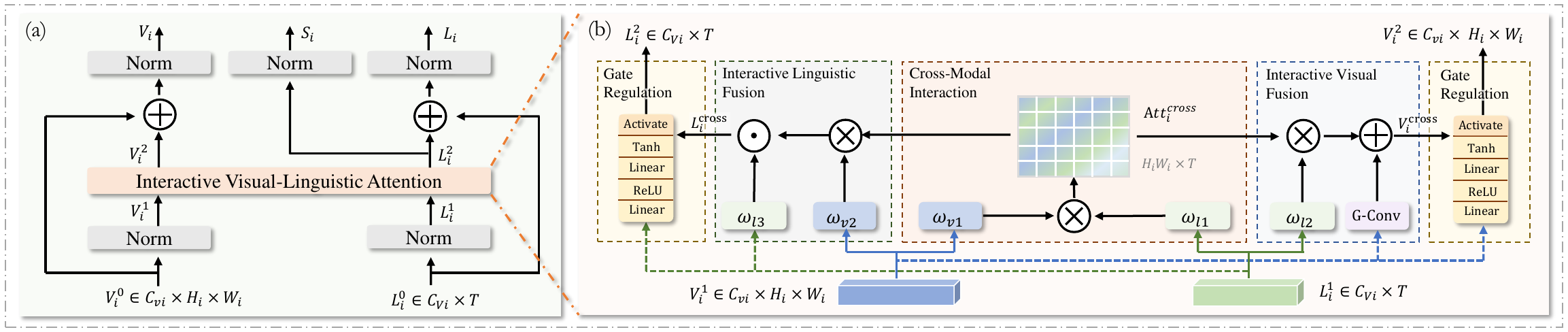}\vspace{-5pt} 
    \caption{(a) An illustration of the interaction block in the Joint Vision-Language Encoder. (b) An illustration of the Interactive Visual-Linguistic Attention.}
    \label{fig3}\vspace{-10pt} 
\end{figure*}

\section{Hierarchical Scale-Aware Vision-Language Transformer}

\subsection{Overview}


The proposed HSVLT simultaneously updates visual, linguistic features and global visual-linguistic relationships within scales as well as performing cross-scale aggregation to aid in multi-label classification prediction.
The overall architecture of HSVLT is presented in Figure~2.

Given an input pair of an image and labels in the dataset, our model generates predictions of object labels that are present within the image.
Our HSVLT adopts a hierarchical multi-scale architecture that follows the workflow of \textbf{\emph{[joint vision-language encoder] - [cross-scale aggregation] - classification}}. HSVLT consists of four stages, each with different numbers of interaction blocks and different feature map resolutions. The encoder (Sec.3.2) includes a novel lightweight attention module (Sec.3.3) that uses interactive cross-modal cues to simultaneously update visual, linguistic features and model global visual-linguistic relationships. We also propose a global cross-scale aggregation module (Sec.3.4) to effectively evaluate multi-modal information across scales and enhance the classification accuracy.
In the following subsections, we describe each components of HSVLT in detail.

\subsection{Joint Vision-Language Encoder}

To enhance the alignment of visual-linguistic features in semantic space, we introduce an joint vision-language encoder that can effectively capture joint visual, linguistic and multi-modal features considering interactive cross-modal cues. Figure~\ref{fig3} illustrates the interaction block structure of our encoder, which incorporates a novel attention mechanism (Sec. 3.3) that replaces the conventional self-attention mechanism. 

As shown in Figure~2(a), our encoder has a pyramid structure, which contains 4 stages with decreasing spatial resolutions. There are $N_i$ interaction blocks in our encoder for the $i$-th stage. 
The visual and linguistic inputs provided into the encoder are denoted as $V_0\in \mathbb{R}^{C_{v0} \times H \times W}$ and $L_0\in \mathbb{R}^{C_{l} \times T}$,
where $H$ and $W$ are height and width of the input image, $T$ is the number of labels, $C_{v0}$ and $C_{l}$ represent the number of channels for the visual and linguistic inputs, respectively.

\paragraph{\textbf{Encoder Workflow}}
Each stage contains a scale transformation block, a channel unification block and a stack of interaction blocks.
Specially in $1$-st stage, we use a word embedding block to extract the linguistic feature $L_1^{0} \in \mathbb{R}^{C_{v1} \times T}$ via a language encoder BERT~\cite{devlin2018bert}, and a patch embedding block to extract the visual feature $V_1^{0} \in \mathbb{R}^{C_{v1} \times H_1 \times W_1}$. 
At the beginning of $2,3,4$-th stage, the visual feature and the linguistic feature pass the scale transformation block and the channel unification block respectively to down-sample the visual feature map and unify the channel dimensions, getting visual feature $V_i^{0} \in \mathbb{R}^{C_{vi} \times H_i \times W_i}$ and linguistic feature $L_i^{0} \in \mathbb{R}^{C_{vi} \times T}$, which are sent to interaction blocks.
For the sake of clarity, in this section, we assume that  $N_i=1$ for all stages. This assumption implies that we are utilizing only one interaction block per stage to illustrate the network.
For each stage, output of joint visual feature $V_i \in \mathbb{R}^{C_{vi} \times H_i \times W_i}$, joint linguistic feature $L_i \in \mathbb{R}^{C_{vi} \times T}$ and joint multi-modal feature $S_i \in \mathbb{R}^{C_{vi} \times T}$ can be obtained as follows:


\begin{equation}
    V_i^{0}, L_i^{0} = \begin{cases}
PE(V_{0}), WE(L_{0}), & i=1 \\
Down(V_{i-1}), Unify(L_{i-1}),& i=2,3,4 \\
\end{cases}
\end{equation}%

\begin{equation}
    V_i, S_i, L_i = Interact(V_i^{0}, L_i^{0}),
\end{equation}

\noindent where $i$ indexes the stage, function $PE(\cdot)$ indicates the patch embedding block, function $WE(\cdot)$ indicates the word embedding block, function $Down(\cdot)$ indicates the scale transformation block, function $Unify(\cdot)$ indicates the channel unification block, function $Interact(\cdot)$ indicates the interaction block.
Both the patch embedding block and the scale transformation block consist of a convolution with stride of $2$ and kernel size of $3 \times 3$, followed by a batch normalization layer. 
The channel unification block employs a $1\times1$ convolution for linear transformation.

\paragraph{\textbf{Interaction Block}} As shown in Figure~3(a), we perform cross-modal interaction in the interaction block as follows:

\begin{equation}
    V_i^2, L_i^2 = IVLA(Norm(V_i^{0}), Norm(L_i^{0})),
\end{equation}

\noindent where function $IVLA(\cdot)$ indicates the Interactive Visual-Linguistic Attention module (Sec.3.3), function $Norm(\cdot)$ indicates the normalization operation. Then we obtain $V_i$, $L_i$ and $S_i$ by 
\begin{math}
    V_i = Norm(V_i^0 + V_i^2),
\end{math}
\begin{math}
    L_i = Norm(L_i^0 + L_i^2),
\end{math}
and
\begin{math}
    S_i = Norm(L_i^2).
\end{math}

\begin{table}[]
\label{table2}
\scriptsize
\centering
\caption{Detailed settings of different stages in our HSVLT.}
\resizebox{0.48\textwidth}{!}{
\renewcommand\arraystretch{1.2}

\begin{tabular}{c|cccc}
\toprule [0.6pt]

            & Stage 1              & Stage 2              & Stage 3              & Stage 4              \\ 
\midrule [0.4pt]
number of blocks ($N_i$)     & 3 & 3 & 27 & 3 \\
\midrule [0.4pt]
visual output size       &    $\frac{H}{2} \times \frac{W}{2}$    &  $\frac{H}{4} \times \frac{W}{4}$   &  $\frac{H}{8} \times \frac{W}{8}$    &   $\frac{H}{16} \times \frac{W}{16}$   \\
\midrule [0.4pt]
channel      &  96       &      192         &     384      &          768   \\

\bottomrule [0.6pt]
\end{tabular}
}
\label{tab1}
\end{table}

\begin{table*}[ht]
\huge
\centering
\label{table1}

\caption{Experimental results on the Pascal VOC 2007 dataset in terms of class-wise precision (AP in \%) and mean average precision (mAP in \%). 
}
\renewcommand{\floatpagefraction}{.9}

\resizebox{\textwidth}{!}{
\renewcommand\arraystretch{1.05}
\begin{tabular}{l|ccccccccccccccccccccc}

\toprule [1.5pt]
Methods & \emph{aero} & \emph{bike} & \emph{bird} & \emph{boat} & \emph{bottle} & \emph{bus} & \emph{car} & \emph{cat} & \emph{chair} & \emph{cow} & \emph{table} & \emph{dog} & \emph{horse} & \emph{motor} & \emph{person} & \emph{plant} & \emph{sheep} & \emph{sofa} & \emph{train} & \multicolumn{1}{c|}{\emph{tv}} & mAP  \\ \hline
CNN-RNN \cite{2016CNN} & 96.7   & 83.1   & 94.2   & 92.8   & 61.2     & 82.1  & 89.1  & 94.2  & 64.2    & 83.6  & 70.0    & 92.4  & 91.7    & 84.2    & 93.7     & 59.8    & 93.2    & 75.3   & 99.7    & \multicolumn{1}{c|}{78.6} & 84.0 \\
RMIC \cite{he2018reinforced}   & 97.1   & 91.3   & 94.2   & 57.1   & 86.7     & 90.7  & 93.1  & 63.3  & 83.3    & 76.4  & 92.8    & 84.4  & 91.6    & 95.1    & 92.3     & 59.7    & 86.0    & 69.5   & 96.4    & \multicolumn{1}{c|}{79.0} & 84.5 \\
RLSD \cite{zhang2018multilabel}   & 96.4   & 92.7   & 93.8   & 94.1   & 71.2     & 92.5  & 94.2  & 95.7  & 74.3    & 90.0  & 74.2    & 95.4  & 96.2    & 92.1    & 97.9     & 66.9    & 93.5    & 73.7   & 97.5    & \multicolumn{1}{c|}{87.6} & 88.5 \\
HCP \cite{wei2015hcp}    & 98.6   & 97.1   & 98.0   & 95.6   & 75.3     & 94.7  & 95.8  & 97.3  & 73.1    & 90.2  & 80.0    & 97.3  & 96.1    & 94.9    & 96.3     & 78.3    & 94.7    & 76.2   & 97.9    & \multicolumn{1}{c|}{91.5} & 90.9 \\
FeV+LV \cite{yang2016exploit} & 97.9   & 97.0   & 96.6   & 94.6   & 73.6     & 93.9  & 96.5  & 95.5  & 73.7    & 90.3  & 82.8    & 95.4  & 97.7    & 95.9    & 98.6     & 77.6    & 88.7    & 78.0   & 98.3    & \multicolumn{1}{c|}{89.0} & 90.6 \\
RDAR \cite{wang2017multi}   & 98.6   & 97.4   & 96.3   & 96.2   & 75.2     & 92.4  & 96.5  & 97.1  & 76.5    & 92.0  & 87.7    & 96.5  & 97.5    & 93.8    & 98.5     & 81.6    & 93.7    & 82.8   & 98.6    & \multicolumn{1}{c|}{89.3} & 91.9 \\
RARL \cite{chen2018recurrent}   & 98.6   & 97.1   & 97.1   & 95.5   & 75.6     & 92.8  & 96.8  & 97.3  & 78.3    & 92.2  & 87.6    & 96.9  & 96.5    & 93.6    & 98.5     & 81.6    & 93.1    & 83.2   & 98.5    & \multicolumn{1}{c|}{89.3} & 92.0 \\
RCP \cite{wang2016beyond}    & 99.3   & 97.6   & 98.0   & 96.4   & 79.3     & 93.8  & 96.6  & 97.1  & 78.0    & 88.7  & 87.1    & 97.1  & 96.3    & 95.4    & 99.1     & 82.1    & 93.6    & 82.2   & 98.4    & \multicolumn{1}{c|}{92.8} & 92.5 \\
SSGRL† \cite{chen2019learning}  & 99.5   & 97.1   & 97.6   & 97.8   & 82.6     & 94.8  & 96.7  & 98.1  & 78.0    & 97.0  & 85.6    & 97.8  & 98.3    & 96.4    & 98.8     & 84.9    & 96.5    & 79.8   & 98.4    & \multicolumn{1}{c|}{92.8} & 93.4 \\
ML-GCN \cite{chen2019multi} & 99.5   & 98.5   & 98.6   & 98.1   & 80.8     & 94.6  & 97.2  & 98.2  & 82.3    & 95.7  & 86.4    & 98.2  & 98.4    & 96.7    & 99.0     & 84.7    & 96.7    & 84.3   & 98.9    & \multicolumn{1}{c|}{93.7} & 94.0 \\
TSGCN \cite{xu2020joint}  & 98.9   & 98.5   & 96.8   & 97.3   & 87.5     & 94.2  & 97.4  & 97.7  & 84.1    & 92.6  & 89.3    & 98.4  & 98.0    & 96.1    & 98.7     & 84.9    & 96.6    & 87.2   & 98.4    & \multicolumn{1}{c|}{93.7} & 94.3 \\
ASL \cite{ridnik2021asymmetric}    & -      & -      & -      & -      & -        & -     & -     & -     & -       & -     & -       & -     & -       & -       & -        & -       & -       & -      & -       & \multicolumn{1}{c|}{-}    & 94.6 \\
CSRA \cite{zhu2021residual}   & 99.9   & 98.4   & 98.1   & 98.9   & 82.2     & 95.3  & 97.8  & 97.9  & 84.6    & 94.8  & 90.8    & 98.1  & 97.6    & 96.2    & 99.1     & 86.4    & 95.9    & 88.3   & 98.9    & \multicolumn{1}{c|}{94.4} & 94.7 \\
MlTr-l \cite{cheng2022mltr}  & -      & -      & -      & -      & -        & -     & -     & -     & -       & -     & -       & -     & -       & -       & -        & -       & -       & -      & -       & \multicolumn{1}{c|}{-}    & 95.8 \\
Q2L \cite{liu2021query2label}     & 99.9   & 98.9   & 99.0   & 98.4   & 87.7     &  \textcolor{blue}{98.6}  & \textcolor{red}{98.8}  & \textcolor{red}{99.1}  & 84.5    & 98.3  & 89.2    & 99.2  & 99.2    & 99.2    & 99.3     & 90.2    & 98.8    & 88.3   & 99.5    & \multicolumn{1}{c|}{95.5} & 96.1 \\
M3TR \cite{zhao2021m3tr}   & 99.9   & 99.3   & 99.1   & \textcolor{red}{99.1}   & 84.0     & 97.6   & 98.0  & \textcolor{blue}{99.0}  & 85.9    & \textcolor{blue}{99.4}  & 93.9    & 99.5  & 99.4    & 98.5    & 99.2     & 90.3    & 99.7    & 91.6   & 99.8    & \multicolumn{1}{c|}{96.0} & 96.5 \\
TSFormer  \cite{zhu2022two}    & 100.0  & 99.2   & 99.2   & 98.6   & 86.4     & 97.2  & 98.4  & 98.9  & 88.9    & \textcolor{red}{99.5}  & 95.3    & \textcolor{red}{99.7}  & 99.6    & 99.1    & 99.4     & 90.0    & \textcolor{blue}{99.6}    & \textcolor{blue}{93.7}   & \textcolor{red}{99.9}    & \multicolumn{1}{c|}{\textcolor{blue}{96.7}} & 97.0 \\ \hline
Ours (w/o CSA) &   \textcolor{blue}{100.0}    &  \textcolor{blue}{99.5}    &  \textcolor{blue}{99.5}    &    \textcolor{blue}{98.7}   &  \textcolor{blue}{89.2}  &  98.3   &  \textcolor{blue}{98.7}  &   98.3  &  \textcolor{blue}{90.6}  &  97.7   & \textcolor{blue}{95.4}  &   \textcolor{blue}{99.3}   &  \textcolor{blue}{99.6}  &  \textcolor{red}{99.6}  &  \textcolor{blue}{99.5}   &  \textcolor{blue}{91.6}  &  \textcolor{red}{99.7}  &   93.1   &  \textcolor{blue}{99.8}   &     \multicolumn{1}{c|}{\textcolor{red}{97.5}} & \textcolor{blue}{97.3}     \\
Ours    & \textcolor{red}{100.0}  &  \textcolor{red}{99.5} &  \textcolor{red}{99.5} & 98.4   & \textcolor{red}{89.4}  &   \textcolor{red}{98.6}  &  98.5 & 98.8  &  \textcolor{red}{93.1}  &  98.4  &  \textcolor{red}{96.2}  &    99.0  &  \textcolor{red}{99.9}  & \textcolor{blue}{99.2}  & \textcolor{red}{99.6}   &  \textcolor{red}{92.5}  &  99.5  &  \textcolor{red}{94.9}  &  99.3  &     \multicolumn{1}{c|}{95.5} & \textcolor{red}{97.5}\\ 

\bottomrule [1.5pt]
\end{tabular}
}
\label{tab2}
\end{table*}

\subsection{Interactive Visual-Linguistic Attention}

As depicted in Figure \ref{fig3}(b), our proposed attention mechanism, namely Interactive Visual-Linguistic Attention (IVLA), 
holistically captures local visual details and global visual-linguistic relationships using interactive cross-modal cues.
IVLA 
comprises four components: a cross-modal interaction to model global visual-linguistic relationships, an interactive linguistic fusion to update the linguistic feature with cross-modal cues, an interactive visual fusion to capture local visual details and incorporates cross-modal cues,
and a gate regulation block to controls the flow of cross-modal knowledge.
In $i$-th stage, given the visual input $V_{i}^1$ and the linguistic input $L_{i}^1$, we obtain visual output $V_{i}^2 \in \mathbb{R}^{C_{vi} \times H_i \times W_i}$ and linguistic output
$L_{i}^2\in \mathbb{R}^{C_{vi} \times T}$ as following steps.

\paragraph{\textbf{Cross-Modal Interaction}}
We utilize a cross-modal interaction to model global visual-linguistic relationships for visual and linguistic features. The steps to get interactive cross-modal activation $Att_i^{cross} \in \mathbb{R}^{H_i W_i \times T}$ are described as the following:

\begin{equation}
    Att_i^{cross} =  \frac{flatten(\omega_{v1} (V_{i}^1) )^T  \omega_{l1} (L_{i}^1)}{\sqrt{C_{vi}}},
\end{equation}%

\noindent where $\omega_{v1}$, $\omega_{l1}$ are projection functions,
and $flatten(\cdot)$ means unrolling the two spatial dimensions into one dimension in row-major.
Here, $Att_i^{cross}$ is the attention scores between $V_{i}^1$ and $L_{i}^1$, which represents the degree of correlation between the two modalities. 
$\omega_{v1}$ is implemented as a 1×1 convolution followed by instance normalization. $\omega_{l1}$ is implemented as a 1×1 convolution. Both $\omega_{v1}$ and $\omega_{l1}$ generate $C_{vi}$ number of output channels.

\paragraph{\textbf{Interactive Linguistic Fusion}}
We activate the linguistic features using interactive cross-modal activation $Att_i^{cross}$, and combine it with the linearly transformed language input through element-wise multiplication, resulting in the cross linguistic feature $L_i^{cross} \in \mathbb{R}^{C_{vi} \times T}$.
$L_i^{cross}$ can be obtained using the following equation:

\begin{equation}
\begin{split}
    L_i^{cross} =\  &  \omega_{l3} (L_{i}^1) \  \odot \\
    & flatten(\omega_{v2} (V_{i}^1))softmax(
    Att_i^{cross}
    ),
\end{split}
\end{equation}%

\noindent where $\omega_{v2}$ and $\omega_{l3}$ are projection functions same as $\omega_{v1}$ and $\omega_{l1}$.

\paragraph{\textbf{Interactive Visual Fusion}}
There is a G-Conv operation to capture local features, which has spatial inductive-bias in modeling rich local visual information. Cross visual feature $V_i^{cross} \in \mathbb{R}^{C_{vi} \times H_i \times W_i}$ can be obtained using the following equation:

\begin{equation}
\begin{split}
    V_i^{cross} =\  & G \text{-} Conv(V_{i}^1) \ + \\ 
    & unflatten((softmax(
    Att_i^{cross}
    )\omega_{l2} (L_{i}^1)^T)^T),
\end{split}
\end{equation}

\noindent where $\omega_{l2}$ indicates projection function same as $\omega_{l1}$, $unflatten(\cdot)$ indicates the opposite operation of $flatten(\cdot)$, and $G\text{-}Conv(\cdot)$ 
denotes the application of a 7×7 convolution operation, followed by a GELU activation function.

\begin{table*}[ht]
\tiny
\centering
\label{table4}

\caption{Experimental results on the Microsoft COCO dataset under the settings of all and top-3 labels (mAP in \%).}
\renewcommand{\floatpagefraction}{.9}

\resizebox{\textwidth}{!}{
\renewcommand\arraystretch{1.05}

{\fontsize{4}{4.8}\selectfont
\begin{tabular}{l|c|cccccc|cccccc}

\toprule [0.5pt]

\multirow{2}{*}{Methods} & \multirow{2}{*}{mAP} & \multicolumn{6}{c|}{All}                                                                                                                         & \multicolumn{6}{c}{Top-3}                                                                                                                        \\ \cline{3-14} 
                         &                      & \multicolumn{1}{c}{CP}   & \multicolumn{1}{c}{CR}   & \multicolumn{1}{c}{CF1}  & \multicolumn{1}{c}{OP}   & \multicolumn{1}{c}{OR}   & OF1  & \multicolumn{1}{c}{CP}   & \multicolumn{1}{c}{CR}   & \multicolumn{1}{c}{CF1}  & \multicolumn{1}{c}{OP}   & \multicolumn{1}{c}{OR}   & OF1  \\ \hline
CNN-RNN  \cite{2016CNN}   & 61.2                 & \multicolumn{1}{c}{-}    & \multicolumn{1}{c}{-}    & \multicolumn{1}{c}{-}    & \multicolumn{1}{c}{-}    & \multicolumn{1}{c}{-}    & -    & \multicolumn{1}{c}{66.0} & \multicolumn{1}{c}{55.6} & \multicolumn{1}{c}{60.4} & \multicolumn{1}{c}{69.2} & \multicolumn{1}{c}{66.4} & 67.8 \\
ResNet-101 \cite{he2016deep}   & 77.3                 & \multicolumn{1}{c}{80.2} & \multicolumn{1}{c}{66.7} & \multicolumn{1}{c}{72.8} & \multicolumn{1}{c}{83.9} & \multicolumn{1}{c}{70.8} & 76.8 & \multicolumn{1}{c}{84.1} & \multicolumn{1}{c}{59.4} & \multicolumn{1}{c}{69.7} & \multicolumn{1}{c}{89.1} & \multicolumn{1}{c}{62.8} & 73.6 \\
SRN  \cite{zhu2017learning}   & 77.1                 & \multicolumn{1}{c}{81.6} & \multicolumn{1}{c}{65.4} & \multicolumn{1}{c}{71.2} & \multicolumn{1}{c}{82.7} & \multicolumn{1}{c}{69.9} & 75.8 & \multicolumn{1}{c}{85.2} & \multicolumn{1}{c}{58.8} & \multicolumn{1}{c}{67.4} & \multicolumn{1}{c}{87.4} & \multicolumn{1}{c}{62.5} & 72.9 \\
ML-GCN  \cite{chen2019multi}  & 83.0                 & \multicolumn{1}{c}{85.1} & \multicolumn{1}{c}{72.0} & \multicolumn{1}{c}{78.0} & \multicolumn{1}{c}{85.8} & \multicolumn{1}{c}{75.4} & 80.3 & \multicolumn{1}{c}{89.2} & \multicolumn{1}{c}{64.1} & \multicolumn{1}{c}{74.6} & \multicolumn{1}{c}{90.5} & \multicolumn{1}{c}{66.5} & 76.7 \\
SSGRL†  \cite{chen2019learning}   & 83.6                 & \multicolumn{1}{c}{89.5} & \multicolumn{1}{c}{68.3} & \multicolumn{1}{c}{76.9} & \multicolumn{1}{c}{\textcolor{red}{91.2}} & \multicolumn{1}{c}{70.7} & 79.3 & \multicolumn{1}{c}{91.9} & \multicolumn{1}{c}{62.1} & \multicolumn{1}{c}{73.0} & \multicolumn{1}{c}{93.6} & \multicolumn{1}{c}{64.2} & 76.0 \\
CMA  \cite{you2020cross}  & 83.4                 & \multicolumn{1}{c}{82.1} & \multicolumn{1}{c}{73.1} & \multicolumn{1}{c}{77.3} & \multicolumn{1}{c}{83.7} & \multicolumn{1}{c}{76.3} & 79.9 & \multicolumn{1}{c}{87.2} & \multicolumn{1}{c}{64.6} & \multicolumn{1}{c}{74.2} & \multicolumn{1}{c}{89.1} & \multicolumn{1}{c}{66.7} & 76.3 \\
KSSNet  \cite{liu2018multi}  & 83.7                 & \multicolumn{1}{c}{84.6} & \multicolumn{1}{c}{73.2} & \multicolumn{1}{c}{77.2} & \multicolumn{1}{c}{87.8} & \multicolumn{1}{c}{76.2} & 81.5 & \multicolumn{1}{c}{-}    & \multicolumn{1}{c}{-}    & \multicolumn{1}{c}{-}    & \multicolumn{1}{c}{-}    & \multicolumn{1}{c}{-}    & -    \\
MCAR \cite{gao2020multi}  & 83.8                 & \multicolumn{1}{c}{85.0} & \multicolumn{1}{c}{72.1} & \multicolumn{1}{c}{78.0} & \multicolumn{1}{c}{88.0} & \multicolumn{1}{c}{73.9} & 80.3 & \multicolumn{1}{c}{88.1} & \multicolumn{1}{c}{65.5} & \multicolumn{1}{c}{75.1} & \multicolumn{1}{c}{91.0} & \multicolumn{1}{c}{66.3} & 76.7 \\
TSGCN  \cite{xu2020joint}  & 83.5                 & \multicolumn{1}{c}{81.5} & \multicolumn{1}{c}{72.3} & \multicolumn{1}{c}{76.7} & \multicolumn{1}{c}{84.9} & \multicolumn{1}{c}{75.3} & 79.8 & \multicolumn{1}{c}{84.1} & \multicolumn{1}{c}{67.1} & \multicolumn{1}{c}{74.6} & \multicolumn{1}{c}{89.5} & \multicolumn{1}{c}{69.3} & 69.3 \\
GM-MLIC \cite{wu2021gm}  & 84.3                 & \multicolumn{1}{c}{87.3} & \multicolumn{1}{c}{70.8} & \multicolumn{1}{c}{78.3} & \multicolumn{1}{c}{88.6} & \multicolumn{1}{c}{74.8} & 80.6 & \multicolumn{1}{c}{90.6} & \multicolumn{1}{c}{67.3} & \multicolumn{1}{c}{74.9} & \multicolumn{1}{c}{\textcolor{blue}{94.0}} & \multicolumn{1}{c}{69.8} & 77.8 \\
C-Tran \cite{lanchantin2021general}   & 85.1                 & \multicolumn{1}{c}{86.3} & \multicolumn{1}{c}{74.3} & \multicolumn{1}{c}{79.9} & \multicolumn{1}{c}{87.7} & \multicolumn{1}{c}{76.5} & 81.7 & \multicolumn{1}{c}{90.1} & \multicolumn{1}{c}{65.7} & \multicolumn{1}{c}{76.0} & \multicolumn{1}{c}{92.1} & \multicolumn{1}{c}{71.4} & 77.6 \\
ADD-GCN  \cite{ye2020attention}    & 85.2                 & \multicolumn{1}{c}{84.7} & \multicolumn{1}{c}{75.9} & \multicolumn{1}{c}{80.1} & \multicolumn{1}{c}{84.9} & \multicolumn{1}{c}{79.4} & 82.0 & \multicolumn{1}{c}{88.8} & \multicolumn{1}{c}{66.2} & \multicolumn{1}{c}{75.8} & \multicolumn{1}{c}{90.3} & \multicolumn{1}{c}{68.5} & 77.9 \\
ASL \cite{ridnik2021asymmetric}   & 86.6                 & \multicolumn{1}{c}{87.2} & \multicolumn{1}{c}{76.4} & \multicolumn{1}{c}{81.4} & \multicolumn{1}{c}{88.2} & \multicolumn{1}{c}{79.2} & 81.8 & \multicolumn{1}{c}{91.8} & \multicolumn{1}{c}{63.4} & \multicolumn{1}{c}{75.1} & \multicolumn{1}{c}{92.9} & \multicolumn{1}{c}{66.4} & 77.4 \\
CSRA   \cite{zhu2021residual}   & 86.9                 & \multicolumn{1}{c}{89.1} & \multicolumn{1}{c}{74.2} & \multicolumn{1}{c}{81.0} & \multicolumn{1}{c}{89.6} & \multicolumn{1}{c}{77.1} & 82.9 & \multicolumn{1}{c}{92.5} & \multicolumn{1}{c}{65.8} & \multicolumn{1}{c}{76.9} & \multicolumn{1}{c}{93.4} & \multicolumn{1}{c}{68.1} & 78.8 \\
Q2L  \cite{liu2021query2label}  & 87.3                 & \multicolumn{1}{c}{87.6} & \multicolumn{1}{c}{76.5} & \multicolumn{1}{c}{81.6} & \multicolumn{1}{c}{88.4} & \multicolumn{1}{c}{78.5} & 83.1 & \multicolumn{1}{c}{91.9} & \multicolumn{1}{c}{66.2} & \multicolumn{1}{c}{77.0} & \multicolumn{1}{c}{93.5} & \multicolumn{1}{c}{67.6} & 78.5 \\
M3TR   \cite{zhao2021m3tr}   & 87.5                 & \multicolumn{1}{c}{88.4} & \multicolumn{1}{c}{77.2} & \multicolumn{1}{c}{82.5} & \multicolumn{1}{c}{88.3} & \multicolumn{1}{c}{79.8} & 83.8 & \multicolumn{1}{c}{91.9} & \multicolumn{1}{c}{68.1} & \multicolumn{1}{c}{78.2} & \multicolumn{1}{c}{92.6} & \multicolumn{1}{c}{69.6} & 79.4 \\
TSFormer  \cite{zhu2022two}   & 88.9                 & \multicolumn{1}{c}{88.3} & \multicolumn{1}{c}{79.2} & \multicolumn{1}{c}{83.5} & \multicolumn{1}{c}{88.5} & \multicolumn{1}{c}{81.5} & 84.9 & \multicolumn{1}{c}{92.3} & \multicolumn{1}{c}{69.1} & \multicolumn{1}{c}{79.0} & \multicolumn{1}{c}{93.2} & \multicolumn{1}{c}{70.5} & 80.3 \\ \hline
Ours (w/o CSA)           & \textcolor{blue}{90.8}                    & \multicolumn{1}{c}{\textcolor{blue}{89.6}}    & \multicolumn{1}{c}{\textcolor{blue}{82.2}}    & \multicolumn{1}{c}{\textcolor{blue}{85.7}}    & \multicolumn{1}{c}{89.1}    & \multicolumn{1}{c}{\textcolor{blue}{84.5}}   & \textcolor{blue}{86.7}    & \multicolumn{1}{c}{\textcolor{blue}{93.0}}    & \multicolumn{1}{c}{\textcolor{blue}{70.7}}    & \multicolumn{1}{c}{\textcolor{blue}{80.3}}    & \multicolumn{1}{c}{94.0}    & \multicolumn{1}{c}{\textcolor{blue}71.7}    & \textcolor{blue}{81.3}    \\
Ours                     & \textcolor{red}{91.6}                   & \multicolumn{1}{c}{\textcolor{red}{89.8}}    & \multicolumn{1}{c}{\textcolor{red}{84.4}}    & \multicolumn{1}{c}{\textcolor{red}{87.0}}    & \multicolumn{1}{c}{\textcolor{blue}{89.8}}    & \multicolumn{1}{c}{\textcolor{red}{86.4}}    & \textcolor{red}{88.0}    & \multicolumn{1}{c}{\textcolor{red}{93.5}}    & \multicolumn{1}{c}{\textcolor{red}{71.6}}    & \multicolumn{1}{c}{\textcolor{red}{81.1}}    & \multicolumn{1}{c}{\textcolor{red}{94.1}}    & \multicolumn{1}{c}{\textcolor{red}{72.6}}    & \textcolor{red}{82.0}    \\ 
\bottomrule [0.5pt]
\end{tabular}}
}
\label{tab3} \vspace{-5pt} 
\end{table*}

\begin{figure}[t!]
    \centering
    \includegraphics[width=1.0\linewidth]{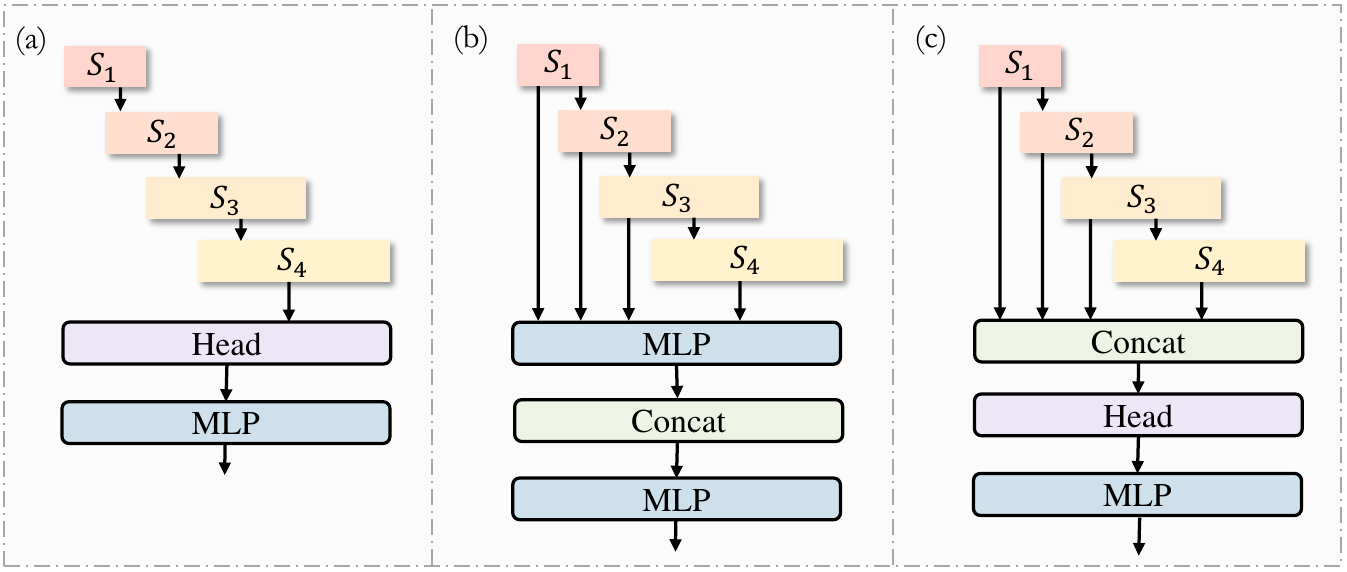}\vspace{-5pt} 
    \caption{Comparison of existing multi-scale integration structures ((a) and (b)) with the structure of our Cross-Scale Aggregation module (c). 
    }
    \label{fig8}\vspace{-20pt} 
\end{figure}

\paragraph{\textbf{Gate Regulation}}
To merge cross-modal knowledge into original visual features $V_i^{1}$ and linguistic features $L_i^{1}$, we introduce Gate Regulation, a gating mechanism. Its core function is to balance the influence of $V_i^{cross}$ and $L_i^{cross}$ on the original information in $V_i^{1}$ and $L_i^{1}$, ensuring controlled passage of cross-modal knowledge to the next stage. A gate unit learns weight mappings from $V_i^{cross}$ and $L_i^{cross}$ to adaptively re-scale each element.
The mathematical formulations of the Gate Regulation is provided below:

\begin{equation}
    V_i^{2} = V_i^{1} + V_i^{cross} \odot Gate(V_i^{cross}),
\end{equation}
\begin{equation}
    L_i^{2} = L_i^{1} + L_i^{cross} \odot Gate(L_i^{cross}),
\end{equation}

\noindent where $Gate(\cdot)$ is composed of two 1×1 convolutions for linear transformation, a ReLU function and a Tanh function.

\subsection{Cross-Scale Aggregation and Classification}

To effectively leverage the visual-linguistic knowledge from different scales, the joint multi-modal features $S_1,S_2,S_3,S_4$ from scales need to be aggregated for final classification prediction. 

We investigate three concise structures shown in Figures~\ref{fig8}(a), \ref{fig8}(b), and \ref{fig8}(c), where \emph{Head} refers to a transformation applied to the channel dimension.
Two typical existing multi-scale integration structures used in other vision tasks are shown in Figures~\ref{fig8}(a), \ref{fig8}(b). The structure adopted by our HSVLT is shown in Figure~\ref{fig8}(c).
Figure~\ref{fig8}(a) is mostly adopted by CNN-based models \cite{zhao2017pyramid, chen2017deeplab}, lacking of holistic consideration to multiple scales. Figures~\ref{fig8}(b) is a purely MLP-based structure \cite{xie2021segformer} with a high computational cost.  As shown in Figure~\ref{fig8}(c), we propose a Cross-Scale Aggregation (CSA) module to aggregate features from the four stages and use a lightweight Hamburger \cite{geng2021attention} to further model the global cross-scale context, resulting in improved performance without compromising computational efficiency.
We obtain the final prediction results by the following equation:

\begin{equation}
    Out = Class(Ham(Concat[S_1, S_2,S_3,S_4])) ,
\end{equation}%

\noindent where $S_i$ is the joint multi-modal feature maps from $i$-th stage, $Ham(\cdot)$ indicates a Hamburger function, and $Class(\cdot)$ indicates a $1 \times 1$ convolution for final prediction.

\begin{figure}[t!]
    \centering
    \includegraphics[width=1.0\linewidth]{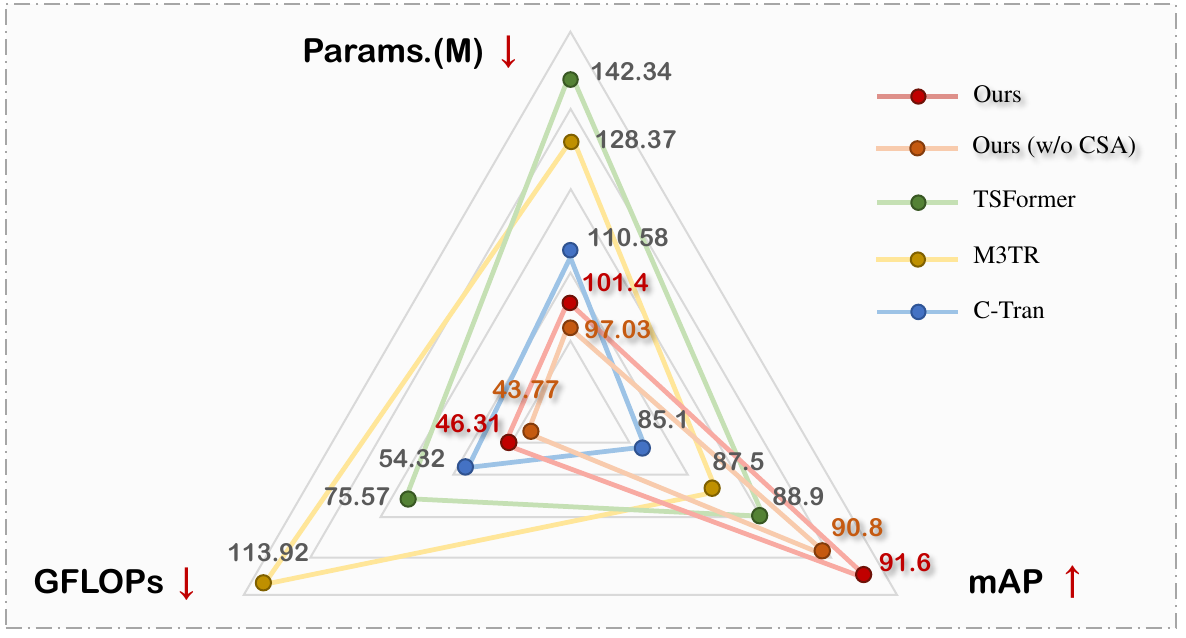}\vspace{-8pt} 
    \caption{ Comparison of our HSVLT (w/o CSA), HSVLT with Transformer-based methods TSFormer \cite{zhu2022two}, M3TR \cite{zhao2021m3tr} and C-Tran \cite{lanchantin2021general} on Params, GFLOPs and mAP on the Microsoft COCO test set. 
    }
    \label{fig6}\vspace{-10pt} 
\end{figure}

\section{Experiments}

\subsection{Dataset and Evaluation}

We perform experiments on three widely used benchmark datasets for multi-label image classification, including Pascal VOC 2007 \cite{everingham2010pascal}, Microsoft COCO \cite{lin2014microsoft}, and NUS-WIDE \cite{chua2009nus}. They have 9,963, 82,081, and 269,648 images respectively, containing 20, 80, and 81 classes. 

Following previous works \cite{zhao2021m3tr,zhu2022two}, we evaluate our proposed method with mean average precision (mAP), a commonly employed metric in the multi-label image classification task. 
To provide a comprehensive comparison, we introduce six additional multi-label metrics, namely overall precision, recall, F1-measure (OP, OR, OF1) and per-class precision, recall, F1-measure (CP, CR, CF1) \cite{chen2019learning}. These metrics are reported under the condition that a label is considered positive if its predicted probability exceeds 0.5. For a systematic comparison with competitors, the top three labels are also reported based on their confidence scores in descending order.

\begin{table}[]
\tiny
\centering

\caption{Experimental results on the NUS-WIDE dataset under the settings of all and top-3 labels (mAP in \%).}

\resizebox{0.48\textwidth}{!}{
\renewcommand\arraystretch{1.05}

\begin{tabular}{l|c|cc|cc}
\toprule [0.57pt]

\multirow{2}{*}{Methods} & \multirow{2}{*}{mAP} & \multicolumn{2}{c|}{All} & \multicolumn{2}{c}{Top-3} \\ \cline{3-6} 
                         &                      & CF1         & OF1        & CF1         & OF1         \\ \hline
CNN-RNN \cite{2016CNN}  & 56.1                 & -           & -          & 34.7        & 55.2        \\
ResNet-101  \cite{he2016deep}   & 59.8                 & 51.9        & 69.5       & 56.8        & 69.1        \\
SRN  \cite{zhu2017learning}  & 62.0                 & 58.5        & 73.4       & 48.9        & 62.2        \\
S-CLs  \cite{liu2018multi}   & 60.1                 & 58.7        & 73.7       & 53.8        & 71.1        \\
CMA  \cite{you2020cross}    & 61.4                 & 60.5        & 73.7       & 55.5        & 70.0        \\
GM-MLIC \cite{wu2021gm}  & 62.2                 & 61.0        & 74.1       & 55.3        & \textcolor{red}{72.5}        \\
ML-GCN  \cite{chen2019multi}  & 62.8                 & 60.7        & 74.1       & 56.3        & 70.6        \\
ASL  \cite{ridnik2021asymmetric}    & 65.2                 & 63.6        & 75.0       & -           & -           \\
MlTr-l  \cite{cheng2022mltr}   & 66.3                 & 65.0        & 75.8       & -           & -           \\
Q2L   \cite{liu2021query2label}  & 66.3                 & 64.0        & 75.0       & -           & -           \\
TSFormer  \cite{zhu2022two}   & 69.3                 & 64.9        & 76.0       & 59.6        & 70.7        \\ \hline
Ours (w/o CSA)                      & \textcolor{blue}{71.4}                    & \textcolor{blue}{66.1}      & \textcolor{blue}{76.4}      & \textcolor{blue}{61.5}     & \textcolor{blue}{72.0}        \\
Ours                     & \textcolor{red}{72.1}                    & \textcolor{red}{66.6}     & \textcolor{red}{76.5}    & \textcolor{red}{61.9}        & 72.0          \\ 

\bottomrule [0.57pt]
\end{tabular}
}
\label{tab4} \vspace{-10pt}
\end{table}

\subsection{Implementation Details}

We conduct experiments using PyTorch library and use
BERT implementation from HuggingFace’s Transformer library \cite{wolf2020transformers}. Following the settings in previous works \cite{zhu2021residual, zhu2022two}, we initialize convolutions in IVLA with weights pre-trained on ImageNet-21K from the ConvNeXt \cite{liu2022convnet}. Language encoder of our model is initialized using official pre-trained weights of BERT with 12 layers and hidden size 768. 
Table~\ref{tab1} presents the detailed network settings, where $N_i$ denotes the number of interaction blocks in the $i$-th stage.
In the interactive visual fusion of IVLA, we use the kernel size of 7×7 for our convolutions.
The rest of weights in our model are randomly initialized.

Following, we use AdamW optimizer with weight decay $0.01$ and batch size $8$. The learning rate is initialed as $1e$-$5$ and scheduled by polynomial learning rate decay with a power of $0.9$. 
To ensure a fair comparison with other models, all input images are adjusted to $448 \times 448$. 
The learning rate is decayed by a factor of 10 when the loss plateaus. 
Following previous works \cite{ridnik2021asymmetric, zhu2022two}, we perform random horizontal flip, random resized crop and RandAugment \cite{cubuk2020randaugment} for data augmentation during the training stage.

\subsection{Comparison with the State-of-the-Arts}

We compare the performance of our proposed HSVLT with state-of-the-art (SOTA) methods on three widely-used benchmarks, namely Pascal VOC 2007 \cite{everingham2010pascal}, Microsoft COCO \cite{lin2014microsoft}, and NUS-WIDE \cite{chua2009nus}. The table below highlights the best scores in \textcolor{red}{red}, and the second-best scores in \textcolor{blue}{blue}, facilitating a straightforward comparison. The symbol † in Tables~\ref{tab2}, \ref{tab3} indicates the utilization of a higher input image resolution (576 × 576).

\paragraph{\textbf{Pascal VOC 2007}}
For equitable assessments, both our model and competitors train on train-val set and test on test set.
Table~\ref{tab2} displays comparative experimental results, including class-wise AP and overall mAP. Our HSVLT achieves SOTA performance in terms of mAP, surpassing other methods. Notably, category AP improvements are evident, e.g., chair, bottle, plant, with HSVLT surpassing 2nd best by 4.2\%, 3.0\%, and 2.5\%. 
Overall, HSVLT achieved the highest AP for more than two-thirds of the label categories.

\begin{table}[]
\small
\centering
\caption{Ablation studies on the Microsoft COCO test set (mAP in \%). The optimal scores are highlighted in bold.}
\resizebox{0.48\textwidth}{!}{
\renewcommand\arraystretch{1.15}

\begin{tabular}{ccccccccc}
\toprule
\multicolumn{1}{l}{} & \multicolumn{1}{l}{} & \multicolumn{1}{l}{} & \multicolumn{1}{l|}{}       & \multicolumn{2}{c}{All}                     & \multicolumn{2}{c|}{Top3}                                  & \multirow{2}{*}{mAP} \\ \cline{5-8}
\multicolumn{1}{l}{} & \multicolumn{1}{l}{} & \multicolumn{1}{l}{} & \multicolumn{1}{c|}{}       & CF1                  & OF1                  & CF1                  & \multicolumn{1}{c|}{OF1}            &                      \\ \hline
\multicolumn{9}{l}{(a) Ablation on kernel size in IVLA}                                                                                                                                                                                    \\ \hline
\multicolumn{4}{c|}{3 × 3} & 85.5 & 86.5 & 79.4 & \multicolumn{1}{c|}{ 80.2 }               & 89.9 \\
\multicolumn{4}{c|}{5 × 5} & 86.1 & 86.9  & 80.3 & \multicolumn{1}{c|}{81.1}               & 90.5 \\
\multicolumn{4}{c|}{7 × 7} &   \textbf{87.0} & \textbf{88.0} & \textbf{81.1}& \multicolumn{1}{c|}{ \textbf{82.0} }               & \textbf{91.6} \\ 
\multicolumn{4}{c|}{11 × 11} & 86.8 & 87.7 & 80.9& \multicolumn{1}{c|}{ 81.7 }               & 91.2 \\ \hline
\multicolumn{9}{l}{(b) Ablation on design choices of IVLA}                                                                                                                                                                                 \\ \hline
G-Conv               & L-Act                & V-Gate               & \multicolumn{1}{c|}{L-Gate} &                      &                      &                      & \multicolumn{1}{c|}{}               &                      \\
\checkmark                    &                      &                      & \multicolumn{1}{c|}{}       & 84.8 & 85.9 & 79.0 & \multicolumn{1}{c|}{ 79.6}               & 89.1 \\
\checkmark                    & \checkmark                    &                      & \multicolumn{1}{c|}{}       & 85.2 & \textbf{88.4} & 79.2 & \multicolumn{1}{c|}{ 80.1 }               & 89.7 \\

\checkmark                    & \checkmark                    & \checkmark                    & \multicolumn{1}{c|}{}       & 86.6 & 87.3 & 80.5 & \multicolumn{1}{c|}{ 81.3 }               & 90.8\\
\checkmark                    & \checkmark                    &                     & \multicolumn{1}{c|}{\checkmark}       & 86.4 & 87.1 & 80.7 & \multicolumn{1}{c|}{ 81.2 }               & 90.7\\

\checkmark                    & \checkmark                    & \checkmark & \multicolumn{1}{c|}{\checkmark}      & \textbf{87.0} & 88.0 & \textbf{81.1}& \multicolumn{1}{c|}{ \textbf{82.0} }               & \textbf{91.6} \\  \hline

\multicolumn{9}{l}{(c) Ablation on design choices of CSA}                                                                                                                                                                                    \\ \hline
\multicolumn{4}{c|}{$S_4$-Head-MLP} &  85.7 & 86.7 & 80.3 & \multicolumn{1}{c|}{ 81.3 }               & 90.8 \\
\multicolumn{4}{c|}{$S_1, S_2, S_3, S_4$-MLP-Concat-MLP} & 86.2 & 87.7 & 80.7 & \multicolumn{1}{c|}{81.5}  &       91.3    \\
\multicolumn{4}{c|}{$S_1, S_2, S_3, S_4$-Concat-Head-MLP} &  \textbf{87.0} & \textbf{88.0} & \textbf{81.1}& \multicolumn{1}{c|}{ \textbf{82.0} }               & \textbf{91.6} \\  \hline
\multicolumn{9}{l}{(d) CSA on various stages} \\ \hline

$S_1$                   & $S_2$                   & $S_3$                   & \multicolumn{1}{c|}{$S_4$}     &                      &                      &                      & \multicolumn{1}{c|}{}               &                      \\
                     &                      &                     & \multicolumn{1}{c|}{\checkmark}      & 85.7 & 86.7 & 80.3 & \multicolumn{1}{c|}{ 81.3 }               & 90.8 \\
            &                      & \checkmark                    & \multicolumn{1}{c|}{\checkmark}      & 86.3 & 87.2 & 80.6 & \multicolumn{1}{c|}{ 81.5 }               & 91.1 \\
\checkmark                    & \checkmark                    & \checkmark & \multicolumn{1}{c|}{ }       & 86.7 & 87.6 & 80.9 & \multicolumn{1}{c|}{81.7}               & 91.3 \\
                     & \checkmark                    & \checkmark                    & \multicolumn{1}{c|}{\checkmark}      & 86.8 & 87.5 & \textbf{81.3} &  \multicolumn{1}{c|}{81.7}               & 91.4 \\
\checkmark                    & \checkmark                    & \checkmark                    & \multicolumn{1}{c|}{\checkmark}      &   \textbf{87.0} & \textbf{88.0} & 81.1& \multicolumn{1}{c|}{ \textbf{82.0} }               & \textbf{91.6} \\ \hline
\multicolumn{9}{l}{(e) Features used for CSA} \\ \hline
\multicolumn{4}{c|}{$L_1, L_2, L_3, L_4$} & 85.3 & 86.6 & 79.9 & \multicolumn{1}{c|}{ 80.7 }               & 90.3 \\
\multicolumn{4}{c|}{$S_1, S_2, S_3, S_4$} &  \textbf{87.0} & \textbf{88.0} & \textbf{81.1}& \multicolumn{1}{c|}{ \textbf{82.0} }               & \textbf{91.6} \\ 
\multicolumn{4}{c|}{$L_1, L_2, L_3, L_4, S_1, S_2, S_3, S_4$} & 86.2 & 87.4 & 80.5 & \multicolumn{1}{c|}{  81.6  }               & 90.9 \\  \hline
\multicolumn{9}{l}{(f) Ablation on different word embedding methods} \\ \hline
\multicolumn{4}{c|}{One-hot \cite{bengio2000neural}} &  86.2  &  87.1 &  80.9  & \multicolumn{1}{c|}{ 81.3   }               & 91.1 \\
\multicolumn{4}{c|}{Glove \cite{pennington2014glove}} &  86.5  &  87.4  &  80.6  & \multicolumn{1}{c|}{ 81.8 }               & 91.3 \\ 
\multicolumn{4}{c|}{BERT \cite{devlin2018bert}} & \textbf{87.0} & \textbf{88.0} & \textbf{81.1}& \multicolumn{1}{c|}{ \textbf{82.0} }               & \textbf{91.6} \\ 
\bottomrule
\end{tabular}

}
\label{tab5} \vspace{-10pt}
\end{table}

\paragraph{\textbf{Microsoft COCO}} 
Following previous studies \cite{chen2019learning, liu2021query2label, zhao2021m3tr}, we report precision, recall, and F1-measure with and without Top-3 scores for the Microsoft COCO dataset, as shown in Table~\ref{tab3}. 
Notably, our HSVLT achieves remarkable mAP of 91.6\%, surpassing all. HSVLT also excels in CF1 and OF1 metrics, in both All and Top-3 contexts.
Furthermore, our HSVLT achieves a higher mAP score while using significantly fewer GFLOPs and parameters compared to previous Transformer-based approaches, as shown in Figure~\ref{fig6}.

\begin{figure*}[ht!]
    \centering
    \includegraphics[width=1.0\linewidth]{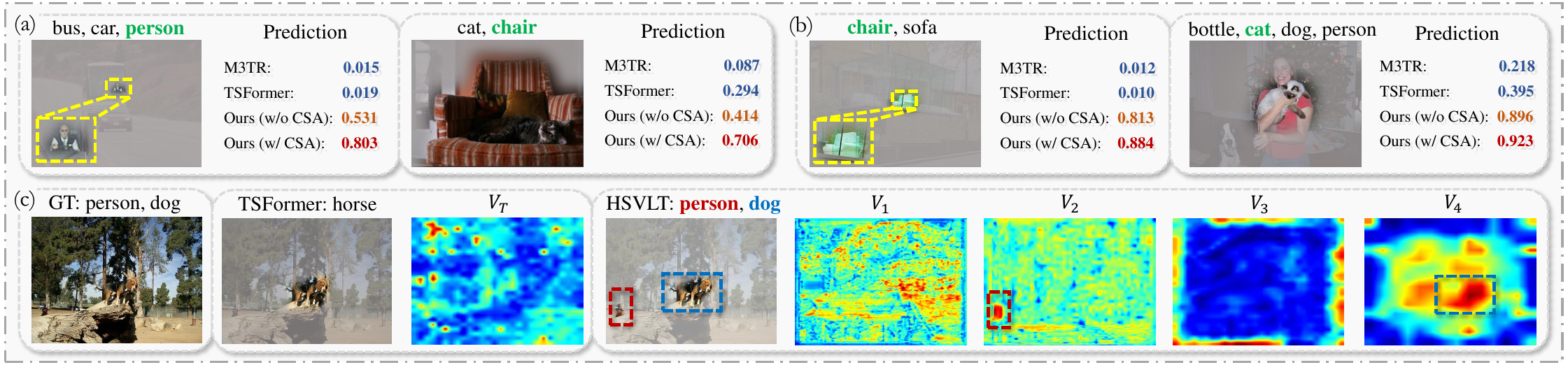}\vspace{-5pt} 
    \caption{(a)~(b)~Comparison of the prediction of a particular label between M3TR \cite{zhao2021m3tr}, TSFormer \cite{zhu2022two} and our HSVLT. For each example, the ground truth label for classification is located above the image, while the predicted values of each model for label in \textcolor{green}{green} are indicated to the right of the image. (c) Comparison of classification results and extracted visual features between TSFormer \cite{zhu2022two} and our HSVLT.  }
    \label{fig9}\vspace{-10pt} 
\end{figure*}

\paragraph{\textbf{NUS-WIDE}}
Test outcomes on NUS-WIDE dataset are in Table~\ref{tab4}. HSVLT excels in critical mAP, surpassing TSFormer by 2.8\%. HSVLT also leads in CF1, OF1 for All, and CF1 for Top-3, showing superiority over SOTA. These findings underscore HSVLT's capability in precise object capture due to its multi-scale structure and joint cross-modal interaction.




\subsection{Ablation Study}

\paragraph{\textbf{Ablation on IVLA design}} We have conducted an ablation study on IVLA design on the Microsoft COCO test set. 
Table~\ref{tab5}(a) presents a comparison of the performance of different convolution kernels in IVLA's interactive visual fusion. The results demonstrate that the 7x7 convolution kernel exhibits superior performance. Moreover, Table~\ref{tab5}(b) details the structure of IVLA and shows the effects of various components, including G-Conv, L-Act, V-Gate, and L-Gate. G-Conv refers to a 7x7 convolution with a GELU activation function, L-Act indicates the dot product activation operation of interactive cross-modal activation $Att_i^{cross}$ on linguistic features, and  V-Gate and L-Gate represent gated units responsible for controlling the flow of the cross-modal knowledge for visual and linguistic features, respectively. The experimental results indicate that the proposed structure achieves the best performance.
It follows that each part contributes to the final performance.

\paragraph{\textbf{Effectiveness of CSA} }
In order to evaluate the effectiveness of the CSA module, we have conducted an ablation study on three benchmark datasets, namely, VOC 2007, Microsoft COCO, and NUS-WIDE. Specifically, we compare the performance of the complete HSVLT with that of HSVLT (w/o CSA) and report the results in Tables~\ref{tab2}, \ref{tab3}, \ref{tab4}, respectively. The experimental results indicate that the removal of the CSA module leads to a degradation in performance, as evidenced by a decrease of 0.2\%, 0.8\%, and 0.7\% in mAP across the three datasets, respectively.
As shown in Table~\ref{tab5}(c), the performance of our proposed CSA outperforms past multi-scale integration structures.
In addition, \emph{Ours} and \emph{Ours (w/o CSA)} in Figure \ref{fig6} also show that CSA improves performance with a slight increase in parameters and GFLOPs, demonstrating the superiority of CSA in interacting multi-modal features among scales.

\paragraph{\textbf{Ablation of CSA on various stages and features}} 
Given joint multi-modal features from different stages, CSA forms features from different scales into a sequence for joint refinement in single forward pass.
$S_i, i \in {1,2,3,4}$ represents multi-modal feature from $i$-th stage inputted to CSA. Table~\ref{tab5}(d) compares multiple input sequences, confirming the value of multi-scale interaction for global reasoning. We also evaluated CSA impact using varied sequences, including $L_i, i \in {1,2,3,4}$ for joint linguistic feature from $i$-th stage. In Table~\ref{tab5}(e), $S_1, S_2, S_3, S_4$ perform optimally for CSA.

\paragraph{\textbf{Ablation of different language models}}
In prior experiments, we used BERT \cite{devlin2018bert} for label semantic embeddings. To assess diverse embedding methods' impact on HSVLT, we contrast it with one-hot encoding \cite{bengio2000neural} and another pre-trained embedding method, namely Glove \cite{pennington2014glove}. The one-hot method encodes labels as one-hot vectors and learns a parameter matrix to map them to the desired embedded space dimension. Table~\ref{tab5}(f) indicates BERT excels, and other methods outperform prior SOTA with same model, confirming HSVLT's adaptability to different language models.


\subsection{Interpretation of HSLVT}

In Figures~\ref{fig9}(a) and \ref{fig9}(b), we compare HSVLT's performance with Transformer-based methods in extreme sizes and confusing appearance predictions. In Figure~\ref{fig9}(a), "person" is much smaller than "bus", and "sofa" largely occupies the image, highlighting HSVLT's object recognition advantage for extreme sizes. In Figure~\ref{fig9}(b), "sofa"-"chair" similarity and "cat"-"dog" confusion are addressed by HSVLT's local visual info and global cross-modal interactions.

Figure~\ref{fig9}(c) contrasts TSFormer and HSVLT in classification results and feature maps. Examining Ground Truth (GT), we observe "person" and "dog" with significant size difference. 
HSVLT's comprehensive consideration of multi-scale information and tighter cross-modal correlations yields precise multi-label results.
$V_T$ and $V_i$ represent features from TSFormer and HSVLT's $i$-th stage. $V_1, V_2, V_3, V_4$ capture diverse aspects of image information. In the example, $V_2$ recognizes the "person", while $V_4$ focuses on the "dog". Thus, the comprehensive aggregation of multi-modal features obtained from different stages helps to enhance multi-label classification performance. 

\section{Conclusion}

In this paper, we propose a novel Transformer-based framework named HSVLT for multi-label image classification. 
HSVLT jointly captures local visual features and models global visual-linguistic relationships considering interactive cross-modal cues at each scale.
The proposed network design interacts multi-modal information between different scales with cross-scale aggregation. 
Experiments show that HSVLT outperforms existing methods on three benchmark datasets with lower computational cost.

\begin{acks}
This work was supported in part by the Major Technological Innovation Project of Hangzhou (No. 2022AIZD0147), National Key Research and Development Project (No. 2022YFC2504605), Zhejiang Provincial Natural Science Foundation of China (No. LZ22F020012), Major Scientific Research Project of Zhejiang Lab (No. 2020ND8AD01).
\end{acks}

\bibliographystyle{ACM-Reference-Format}
\bibliography{sample-sigconf}

\appendix

\end{document}